\documentclass[letterpaper, 10 pt, conference]{ieeeconf}

\IEEEoverridecommandlockouts
\usepackage{array}
\usepackage{booktabs}  
\usepackage{tabularx}  
\usepackage[T1]{fontenc}
\usepackage{cite}
\usepackage{amsmath,amssymb,amsfonts}
\usepackage{algorithmic}
\usepackage{graphicx}
\usepackage{textcomp}
\usepackage{xcolor}
\usepackage{caption}
\usepackage{algorithm,algorithmic}
\usepackage{lipsum}
\usepackage{gensymb}
\usepackage{mleftright}
\usepackage{tikz}
\usetikzlibrary{shapes.geometric, arrows.meta, positioning}
\usepackage{bm}
\usepackage{soul}
\usepackage{comment}
\usepackage{mathrsfs}
\usepackage{multirow}
\usepackage{tikz}
\usepackage{algorithm}
\usepackage{algorithmic}
\setstcolor{red}
\usepackage{makecell}
\let\labelindent\relax
\usepackage{enumitem}
\usepackage{cancel}
\usepackage{subcaption}

\usepackage{amsthm}
\usepackage{hyperref}
\usepackage[algo2e,ruled,vlined,linesnumbered,resetcount,norelsize]{algorithm2e}
\renewcommand{\baselinestretch}{.98}
\renewcommand{\arraystretch}{1.2}

\makeatletter
\newcommand{\linebreakand}{%
  \end{@IEEEauthorhalign}
  \hfill\mbox{}\par
  \mbox{}\hfill\begin{@IEEEauthorhalign}
}
\makeatother

\def\BibTeX{{\rm B\kern-.05em{\sc i\kern-.025em b}\kern-.08em
    T\kern-.1667em\lower.7ex\hbox{E}\kern-.125emX}}
\begin{document}


\title{Mutual Adaptation in Human-Robot Co-Transportation with Human Preference Uncertainty\\
}
\author{
Al Jaber Mahmud, Weizi Li, and Xuan Wang
\thanks{A. Mahmud and X. Wang are with George Mason University. W. Li is with the University of California, Riverside. Supplementary material of this work can be found at \cite{supp}.} 
}

\maketitle
\begin{abstract}

Mutual adaptation can enhance overall task performance in human-robot co-transportation by integrating both the robot's and the human's understanding of the environment. While human modeling helps capture humans' subjective preferences, two challenges persist: (i) the uncertainty of human preference parameters and (ii) the need to balance adaptation strategies that benefit both humans and robots. In this paper, we propose a unified framework to address these challenges and improve task performance through mutual adaptation. First, instead of relying on fixed parameters, we model a probability distribution of human choices by incorporating a range of uncertain human preference parameters. Building on this, we introduce a time-varying stubbornness measure and a coordinated planning model, which allows either the robot to lead the team’s trajectory or, if a human's preferred path conflicts with the robot's plan and their stubbornness exceeds a threshold, the robot to transition to following the human. Finally, we introduce a pose optimization strategy for low-level control to mitigate the uncertain human behaviors when they are leading. To validate the framework, we design and perform a study with human feedback from twenty human participants. We then demonstrate, through simulations, the effectiveness of our models in enhancing task performance with mutual adaptation and pose optimization.


\end{abstract}

\section{Introduction}

Human-robot collaboration (HRC) has gained increasing attention in research due to its potential to enhance efficiency and safety in daily tasks in which robots operate in close proximity to humans. Seamless cooperation requires robots to understand human behavior and anticipate human decision-making uncertainties.
Classical cognitive and behavioral models offer theoretical frameworks to capture human biases, heuristics, and risk attitudes~\cite{tversky1992advances, stevens1963subjective, prelec1998probability}. However, the effectiveness of these models depends on accurate identification of model parameters, which is often challenging because they may vary across individuals and fluctuate even for the same individual under different experimental conditions. On the other hand, data-driven and learning-based approaches~\cite{fintz2022using, gorgan2024computational} have demonstrated success in modeling human behavior, but often lack
explainability and generalizability. Despite extensive research on human behavior prediction, only a few studies~\cite{
sadrfaridpour2016modeling, 
nikolaidis2017human, LP-NT-DC-AA:17} integrate human behavior modeling into low-level robot control during collaboration. 
However, these works typically focus on static, one-time interactions rather than dynamic, evolving environments.

\begin{figure}
    \centering
    \includegraphics[width=0.44\textwidth]{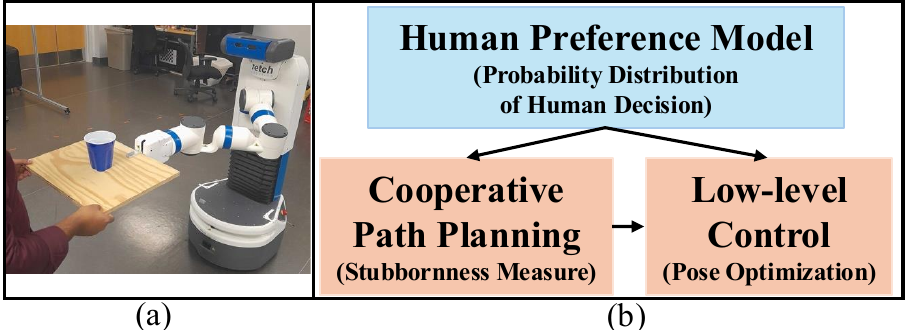}
    \caption{\small (a) Task: Human-Robot Co-Transportation. 
    (b) Overview: Our human preference model informs cooperative path planning and low-level control of the robot. 
    \vspace{-2.2em}
    }
    \label{environment}
\end{figure} 

In this work, we consider a co-transportation task (Fig.~\ref{environment}), where a human-robot team navigates from a start position to a target through obstacle \textit{openings}. Throughout this process, variations in human preferences can significantly influence path selection. If the robot fails to account for these preferences, task performance may degrade, potentially leading to failure.
We study this problem through three hypotheses (Sec.~\ref{sec:problem_interest}) that connect the human preference model with parameter uncertainty, cooperative path planning, and low-level robot control.
Our main contributions are:

\begin{itemize}[leftmargin=*]
    \item \textbf{Human preference model with parameter uncertainty.} We build a mathematical model of human preferences that captures uncertainties in human parameters regarding their perceptions of risk and distance across multiple options. The model generates probability distributions that represent the likelihood of human decisions on each choice, and this human model informs both cooperative path planning and low-level control.
    
    \item\textbf{Cooperative path planning through a stubbornness measure.} We introduce a time-varying stubbornness measure, integrated with probability distributions of human decisions into a coordination mode transition model for cooperative team decision-making. This enables both the human and the robot to adapt to each other dynamically over time.

    \item \textbf{Low-level control through pose optimization}. We develop a pose optimization strategy that accounts for potential deviations in human behavior, allowing the robot to proactively mitigate the impact of unexpected changes between path options in human decisions while ensuring safe and efficient object transportation. The controller uses the probability distributions of human decisions to compensate for potential switching between path options.
\end{itemize}

\section{Literature Review}


\subsubsection*{Human Behavior Model}

Understanding human behavior is crucial for efficient and safe human-robot collaboration. Various methods, including deep learning~\cite{fintz2022using}, reinforcement learning~\cite{tomov2021multi}, and spiking neural networks~\cite{gorgan2024computational} have been employed to learn human behaviors from data. However, these data-driven approaches often lack interpretable mathematical frameworks that describe human decision-making processes.
Classical cognitive and behavioral models, such as prospect theory~\cite{tversky1992advances}, Prelec's probability weighting function~\cite{prelec1998probability}, Stevens' power law~\cite{ stevens1963subjective}, drift-diffusion models~\cite{ratcliff2008diffusion}, and Bayesian models~\cite{eguiluz2015bayesian}, provide theoretical structures to capture human perception and decision-making. Yet, these models often yield specific human parameter values that may not generalize well across individuals~\cite{ prabhudesai2023understanding}. 
Moreover, humans may resist adapting to a robot if they perceive continuous disagreement, despite initial willingness to adapt~\cite{heskes2000use}.


\subsubsection*{Mutual Adaptation}
To analyze the adaptation modes in human-robot collaboration, many studies have focused on \emph{single-direction robot adaptation} to human actions. Such approaches include co-transportation~\cite{sirintuna2023object},
co-manipulation tasks~\cite{noohi2016model},
robot learning from demonstrations~\cite{akgun2012trajectories}, adaptive motion planning under uncertainties~\cite{kanazawa2019adaptive}, 
human-robot handovers~\cite{wu2019line},
and adaptive task planning in shared environments~\cite{bolu2021adaptive}. 
Similarly, there exist \emph{single-direction human adaptation} methods, where the robot leads while the human adapts, such as in robotic tour-guiding~\cite{vasquez2020tour} and complex navigation assistance~\cite{cai2024navigating}.
Although single-direction adaptation is sufficient for some tasks, it may cause human discomfort or disengagement~\cite{van2020adaptive}, riskier task approaches, or longer execution times~\cite{kwon2020humans}. Such drawbacks highlight the need for \emph{mutual adaptation} between the human and the robot for enhanced efficiency in collaboration~\cite{nikolaidis2017human, kumar2024sequential}, in particular, by incorporating human modeling techniques mentioned above.


\subsubsection*{Pose Optimization}

Despite incorporating human variability and behavioral change, mathematical models cannot perfectly capture every individual’s behavior and decision-making processes~\cite{hsu2005neural, prabhudesai2023understanding}. 
Consequently, robots must be prepared to handle uncertainties and unexpected human choices. Although previous studies have discussed human uncertainty in decision-making~\cite{bland2012different, 
johnson2010decision}, adapting robots at low-level control to unexpected human choices remains a challenge.
For such low-level control in the robot's end-effector, the idea of pose optimization, which is related to redundancy resolution~\cite{cavacanti2017redundancy
}, helps to avoid kinematic singularities~\cite{LH-LB:08}, improving the overall reachability of the robot manipulator~\cite{HC-ME-TD:21}. 
A similar strategy has been explored in~\cite{11246189};
however, it did not explicitly integrate a human behavioral model with a probabilistic distribution of human choices. Hence, we need a unified approach that combines human modeling with uncertain human parameters, the mutual adaptation among humans and robots, and pose optimization to ensure safe and effective collaboration.


\section{Preliminaries and Problem Formulation}

\subsection{The Co-transportation Task and Path Selection}
In this paper, we consider a co-transportation task in which a human and a robot jointly transport an object from a start position to a target position. 
The environment may contain multiple obstacles that must be avoided to ensure navigation safety. We assume that both the human and the robot have complete knowledge of the environment. 
Depending on the geometric properties of these obstacles, the human-robot team will have to navigate through the `openings' among obstacles, leading to multiple path options. 
Assuming the considered paths do not involve loops, the total number of path options is finite. 
To evaluate feasible paths for the co-transportation task, we consider two factors: the collision risks $S$ when transporting through all openings and the total traveling distances $D$. 
We evaluate these quantities for the remaining path from the human-robot team’s current position to the target.
Specifically, let $\Xi$ be the set of all openings. An opening $\xi\in \Xi$ is defined as a passage between obstacles whose width satisfies $d_r<w_{\xi}\le 5d_r$,\footnote{When a passage is wider than $5d_r$, we assume the human-robot team can traverse it without incurring any risk. This threshold may be adjusted based on real-world scenarios.} where $d_r$ is the width of the robot. The risk associated with opening $\xi$ is quantified as $S_{\xi} = \frac{d_r}{w_{\xi}}$.
Here $S_{\xi}\in[0,1]$ is the percentage of collision risk for passing through opening $\xi$. In general, a narrower passage implies a higher risk.

A feasible path option $o\in\mathcal{O}$ connects the system’s current position to the target point through a sequence of openings. Here, $\mathcal{O}$ denotes the set of all path options.\footnote{Note that $\mathcal{O}$ may change dynamically, depending on the current positions of the human-robot team. Here, we introduce the generic concept of cost evaluation and omit time-step indicators.}
Let $\Xi_o\subset \Xi$ be the set of openings associated with option $o$, and the percentage of collision risk $S_o$ for this option is then defined by the stacking of risk as $S_o=1 -  \prod_{\xi\in\Xi_o} (1-S_\xi)$.
Let $D_o$ represent the total length of path option $o$ to the target. The overall cost for option $o$ is quantified as 
\begin{align}
    \mathcal{J}_{o} =  D_{o} + c_{\text{r}} \cdot S_{o},
    \label{eq:robot_utility}
\end{align}
where $c_{\text{r}}$ is a scalar weight that determines the relative importance of collision risk versus travel distance. We use the same $c_{\text{r}}$ across options to keep risk-distance trade-offs similar.
Note that \eqref{eq:robot_utility} represents an \textbf{objective} cost that can be utilized by robots to optimize their performance by balancing between safety (risk minimization) and efficiency (distance minimization). 
Nevertheless, in most cases, humans perceive risks and distances subjectively, leading to a discrepancy between human and robotic evaluations, as we introduce next.

\subsection{Subjective Human Cost with Risk-Distance Perception}

\begin{figure}[t]
    \centering
    \includegraphics[width=0.44\textwidth]{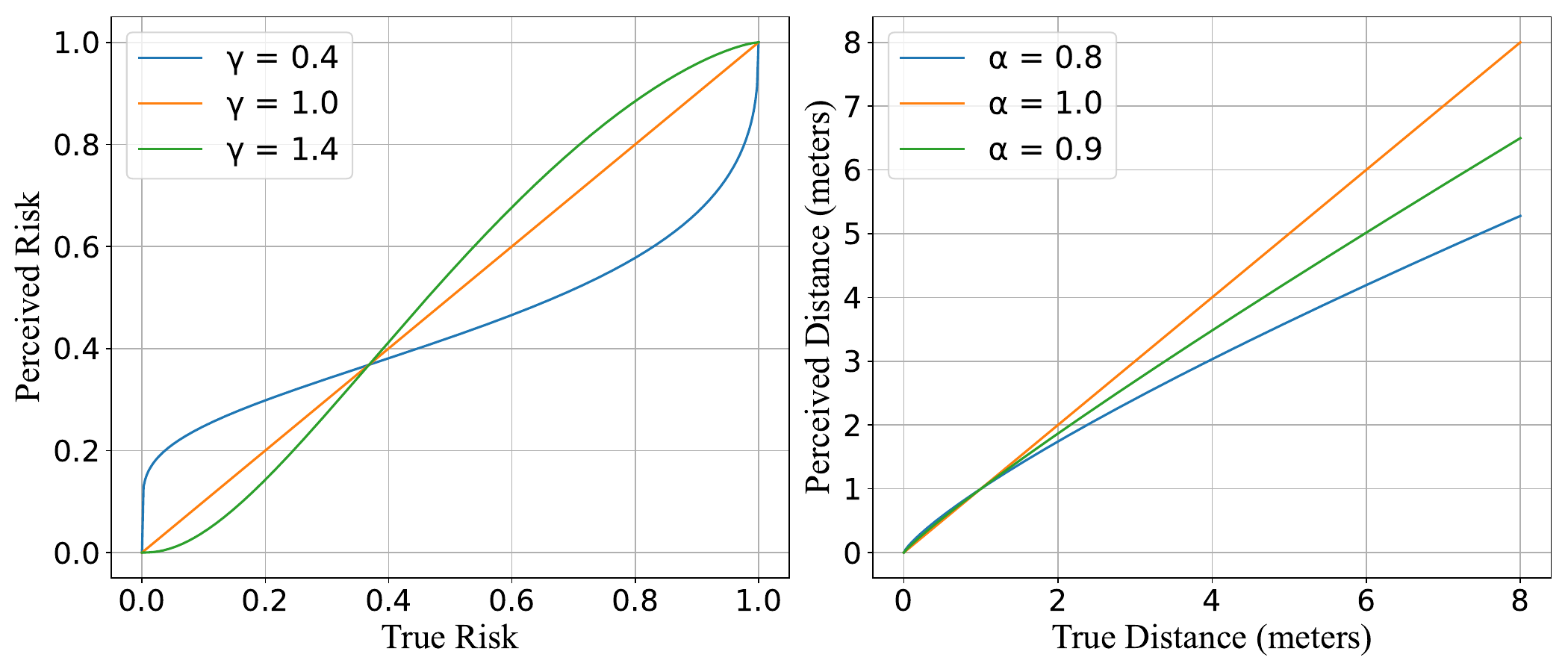}
    \caption{ \small
    True risk $(\gamma=1)$ and true distance $(\alpha = 1)$. For perceived risk, when 
    $\gamma<1$ the weighting function overemphasizes small risks and underemphasizes large risks, whereas when 
    $\gamma>1$ the opposite trend is observed. Perceived distance varies with $\alpha$, reflecting differences in individual sensitivity.
    \vspace{-2.2em}
    }
    \label{perceived_risk_distance}
\end{figure}

Humans tend to exhibit nonlinear risk perception, often overestimating small risks while underestimating large ones~\cite{tversky1992advances}. To capture this phenomenon, we introduce a subjective risk measure $\widehat{S}_{o}$ based on Prelec’s function~\cite{prelec1998probability} and quantified as $\widehat{S}_{o}=   \exp \Bigl( - \bigl( - \ln (S_o) \bigr)^\gamma \Bigr)$,
where $\gamma$ is a \emph{risk-perception} parameter. It controls how a human subjectively weights the objective risk probability $S_o$ to the perceived risk $\widehat{S}_o$. When $\gamma=1$, the perceived risk matches the true risk ($\widehat{S}_o = S_o$). When $\gamma<1$, humans overweight small risk probabilities and underweight large risk probabilities. For $\gamma>1$, the perception towards probabilities is the opposite (Fig.~\ref{perceived_risk_distance}).
Similarly, humans may not perceive the actual travel distance accurately. Inspired by Stevens' power law~\cite{stevens1957psychophysical, stevens1963subjective, thomas2023psychophysics}, 
we model an individual’s sensitivity to distance as $\widehat{D}_{o}=D_{o}^\alpha$,
where $\alpha$ is a \emph{distance-perception} parameter. It controls how a human subjectively weights the true path length $D_o$ to form the perceived distance $\widehat{D}_o$. When $\alpha=1$, perceived distance matches the true distance ($\widehat{D}_{o}=D_{o}$). When $\alpha<1$ (often observed), the perceived distance grows slowly and thus, longer paths feel less costly relative to their actual length (Fig.~\ref{perceived_risk_distance}).
Combining these elements, the human’s \textbf{subjective} cost for an option $o$ is defined as
\vspace{-0.2cm}
\begin{align}\label{eq_subjc}
    \widehat{\mathcal{J}}_{o}(\alpha,\gamma)
    = \widehat{D}_{o} + c_{\text{r}} \cdot \widehat{S}_{o}.
\end{align}

\vspace{-0.1cm}
\noindent
This function describes how humans form subjective judgments based on their own attitude toward risk (whether risk-seeking or risk-averse, governed by \(\gamma\)) and their ability to measure distance through observations (governed by \(\alpha\)).
Although this model can be used to predict human decisions, 
setting the parameters with specific values is not guaranteed to be accurate, as they vary among individuals and even for the same individual under different experimental conditions. Thus, the uncertainties in these parameters must be taken into account.

\subsection{Problem of Interest and Hypotheses}
\label{sec:problem_interest}

In this paper, we investigate the human-robot co-transportation task while explicitly considering uncertainty in subjective preference parameters and time-varying human stubbornness. 
We study this problem through the following hypotheses.

\noindent\textbf{H1 (Preference uncertainty).} If human preference parameters are uncertain rather than fixed, then a probabilistic model over options can better represent how humans distribute their choices across multiple paths. 

{\noindent\textbf{H2 (Mutual adaptation via mode transition).} If the robot uses a time-varying stubbornness measure to decide when the human would take the lead, then the team can achieve lower overall task cost than a fixed-mode baseline where the robot always leads. }

{\noindent\textbf{H3 (Pose preparation under choice uncertainty).} If the robot performs pose optimization (change in joint angle configuration while keeping the same end-effector pose, when necessary) in human-leading mode to prepare for multiple possible human choices, then the robot can reduce low-level control cost compared to not using pose optimization.}

To test these hypotheses, we address three research questions and evaluate them in Sec.~\ref{sec_results}.
(i) \emph{Human modeling:} How to quantify human decision-making uncertainties that arise from variations in their subjective preference parameters (\textbf{H1}).
(ii) \emph{Human-robot alignment:} Given the human preference uncertainties, how does the human shift from follower to leader role using a time-varying stubbornness measure to enable mutual adaptation (\textbf{H2}).
(iii) \emph{Robot control:} How the human model and human-robot alignment mechanism can inform the low-level control of the robot to mitigate the effect of choice uncertainty in human-led mode (\textbf{H3}).

\section{Main Approach}

\begin{figure}
    \centering
    \includegraphics[width=0.49\textwidth]{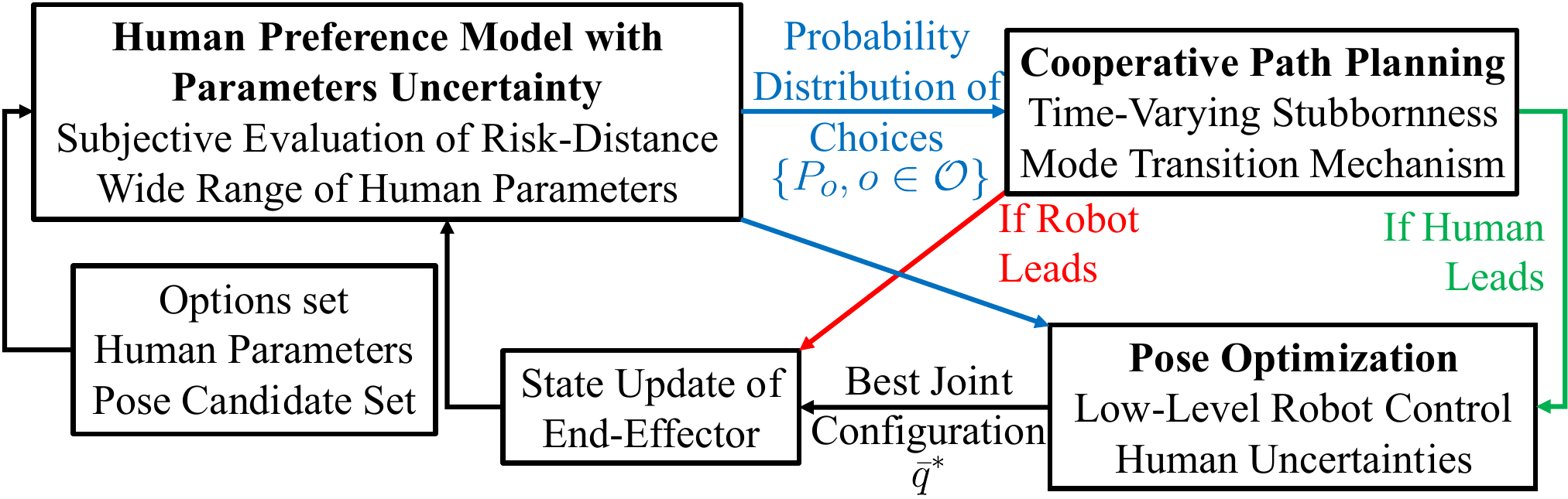}
    \caption{\small Overview of our proposed mutual adaptation framework. The \emph{Human Preference Model with Parameters Uncertainty} produces a probability distribution of options. 
    A time-varying human stubbornness measure is developed that incorporates these probabilities into a \emph{Cooperative Path Planning Model} to determine their coordination modes. If human leads, the \emph{Pose Optimization}, informed by the probability distribution, generates joint configurations that better prepare the robot's low-level control to handle potential human uncertain behaviors.
    \vspace{-2.2em}
    }
    \label{flowchart}
\end{figure}

Aligned with our problem formulation to account for human preference uncertainties, our approach to the human-robot co-transportation task comprises three key components, each corresponding to one hypothesis, mentioned in Sec.~\ref{sec:problem_interest}. 
(i) A human preference model that computes a probability distribution representing the likelihood of a human selecting a particular path option. {This addresses \textbf{H1} by converting parameter uncertainty into a choice distribution over options.} 
(ii) A measure of human stubbornness that informs the mode of human-robot coordination and guides the robot in selecting paths that accommodate human preferences. {This addresses \textbf{H2} by using a time-varying stubbornness level for transitioning from robot-led to human-led mode.}
(iii) A pose optimization strategy integrated into the robot’s low-level control system, which proactively compensates for uncertainties in human decision-making while the robot follows the human. {This addresses \textbf{H3} by preparing robot joint configurations under intent uncertainty during human-led mode.} 
These modules are interconnected, as illustrated by the flowchart in Fig.~\ref{flowchart}.

\subsection{Probability Distribution of Human Choices} \label{sec:HPM}

In this subsection, we develop a mathematical model that maps uncertainty in human decision-making parameters into a probability distribution representing the likelihood of a human choosing a particular option.
Based on the subjective cost defined in \eqref{eq_subjc}, we assume the human chooses the option with the lowest subjective cost, i.e.,
\vspace{-0.15cm}
\begin{align}\label{eq_H_decision}
    o_h(\alpha,\gamma) = \arg\min_{o\in\mathcal{O}} ~~~\widehat{\mathcal{J}}_{o}(\alpha,\gamma)
\end{align}

\vspace{-0.1cm}
\noindent
where $o_h (\cdot)$ is the human preferred option, which depends on the parameters $\alpha$ and $\gamma$.
{We do not assume the human always executes this option during co-transportation. The human may follow the robot at the beginning and then may lead the team through mode transition (Sec.~\ref{HBM}). Eq.~\eqref{eq_H_decision} is primarily used to compute the probability distribution over options $\{P_o\}$ under preference parameter uncertainty. This distribution then contributes to (i) accumulation of human discomfort and mode transition (Sec.~\ref{HBM}), and (ii) pose selection under human intent uncertainty (Sec.~\ref{PO}).}

\begin{figure}
    \centering
    \includegraphics[width=0.25\textwidth]{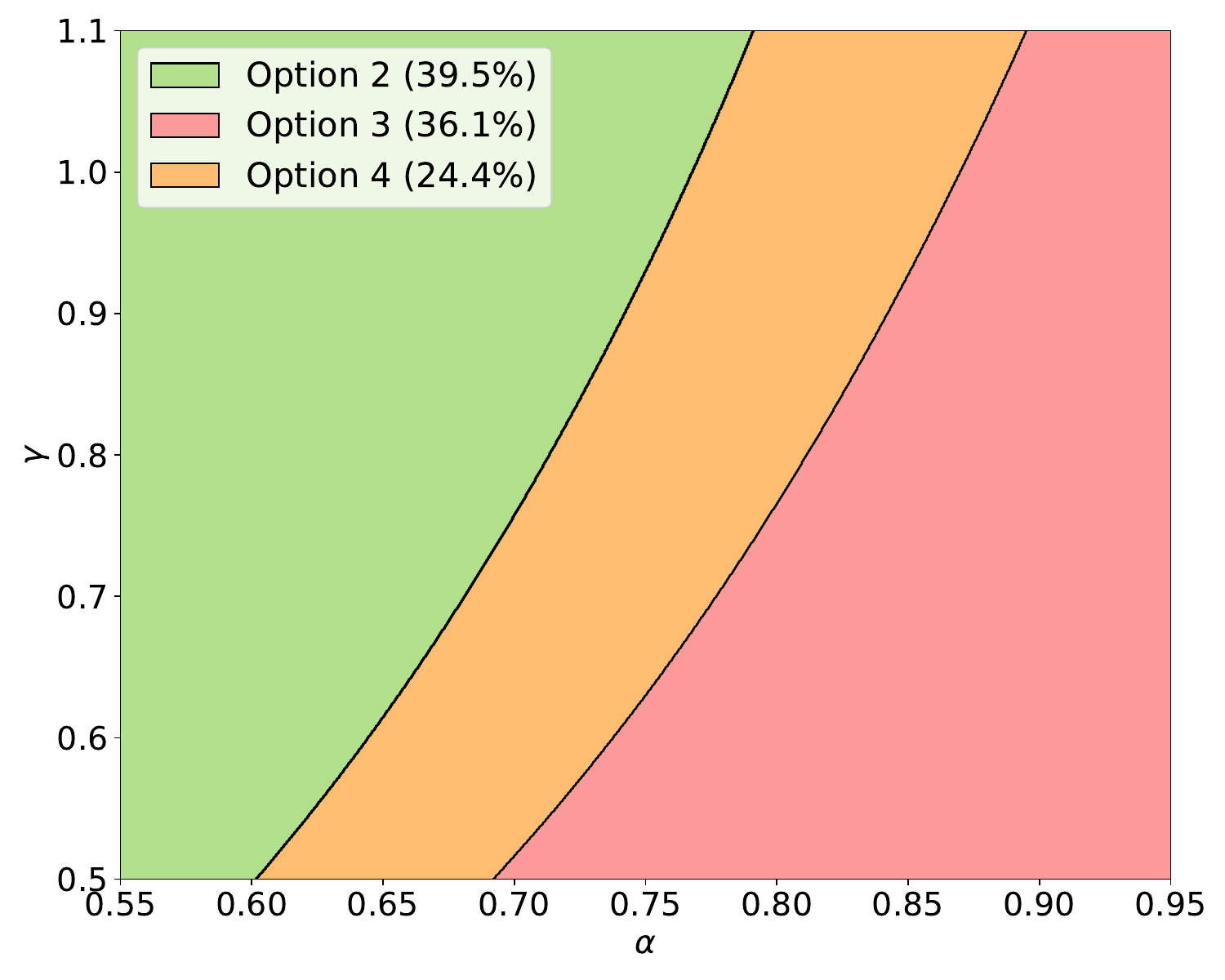}
    \caption{\small With the environment in Fig.~\ref{environment}, the probability distribution of choosing an option based on the parameter distributions $\alpha\sim\mathrm{U}(0.55, 0.95)$ and $\gamma\sim\mathrm{U}(0.5, 1.1)$ with $c_\text{r} = 10$.
    The colored regions correspond to the lowest subjective cost of each option. The probabilities are computed based on the areas of these regions. 
    \vspace{-3.2em}
    }
    \label{probability_distribution}
\end{figure}

To account for uncertainties of human preference parameters, we model $\alpha$ and $\gamma$ as two independent variables following a uniform distribution\footnote{The results can be easily generalized to other types of distributions, such as the normal distribution.} such that $\alpha \sim \mathrm{U}(\underline{\alpha}, \overline{\alpha})$ and $\gamma \sim \mathrm{U}(\underline{\gamma}, \overline{\gamma})$, 
where $\underline{.}$ and $\overline{.}$ represent the lower and upper bound of the distribution, respectively.
Hence, for each pair of $(\alpha,\gamma)$, the human's preferred option $o_h(\alpha,\gamma)$ may differ, leading to a distribution over the preferred options.
As illustrated in Fig.~\ref{probability_distribution}, this partitions the \((\alpha, \gamma)\) plane into regions \(\mathcal{R}_{\mathcal{O}}\) defined as $\mathcal{R}_{o_h} \;=\; \Bigl\{ (\alpha, \gamma)
\;\Big|\;
o_h
\Bigr\}.$
In particular, {$\mathcal{R}_{o}$ is the set of parameter pairs for which option $o \in \mathcal{O}$ achieves the lowest subjective cost in \eqref{eq_H_decision}.}
Given the uniform distributions for \(\alpha\) and \(\gamma\), the probability that the human chooses option \(o\) is formulated as the ratio of the area of \(\mathcal{R}_{\mathcal{O}}\) to the total area of the parameter space:
\vspace{-0.15cm}
\begin{align}\label{eq_Po}
    P_{(o=o_h)} 
    \;=\; 
    \frac{1}{(\overline{\alpha} - \underline{\alpha})(\overline{\gamma} - \underline{\gamma})}
    \iint_{\mathcal{R}_{o_h} }
    d\alpha \, d\gamma.
\end{align}
Thus, our model yields a probability distribution over the set of options, \( \{P_o,o\in\mathcal{O}\} \), allowing the robot to anticipate variability in human preferences.
In this paper, we consider $\alpha\sim\mathrm{U}(0.55, 0.95)$ and $\gamma\sim\mathrm{U}(0.5, 1.1)$.
The range of these values aligns with the extensive human experiments in existing research~\cite{tversky1992advances,prelec1998probability, abdellaoui2007loss, thomas2023psychophysics}. Parameter $c_\text{r}$ is task-specific and will be identified in our experiments using real human feedback.

\subsection{Time-Varying Human Stubbornness in Human-robot Coordination} \label{HBM}

Effective human-robot collaboration requires balancing the robot’s objective evaluation with the human’s preference to achieve a team decision that enables mutual adaptation. In the context of co-transportation, we consider two modes: either the human follows the robot’s guidance or the robot follows the human. 
To model the mode transition, we first introduce a time-varying measure of human stubbornness that quantifies how resistant a human is to deviating from their preferred path option. Then we incorporate it into a mode transition model. When the robot is leading the team, the model also helps to determine which path option the robot should choose that accounts for both the objective cost of the robot and the subjective cost of the human.

\subsubsection{Modeling Human Stubbornness}
We define a time-varying stubbornness measure \(\phi(k)\) as follows, where $k$ denotes the time step,
\vspace{-0.3cm}
\begin{align}
    \phi(k) =  \phi(0) \cdot \exp\Bigl(\frac{g(k)}{\eta}\Bigr)
    \label{eq:phi_evolution}
\end{align}

\vspace{-0.4cm}
\noindent
where, 

\vspace{-0.6cm}
\begin{align*}
    g(k) =  g(k-1)+ \sum_{o\in \{\mathcal{O}\}} \omega_o(k) P_o (k),~~~g(0)=0.
\end{align*}

\vspace{-0.25cm}
\noindent Here, $g(k)$ approximates the accumulation of human discomfort. It grows along the trajectory if the team’s movement does not align with human preferences. The accumulation coefficient $\omega_o(k)\in\{0,1\}$ is defined such that $\omega_o(k)=1$ (not aligned) if the team moves towards an opening $\xi$, that is \textbf{not} the next waypoint in option $o$; $\omega_o(k)=0$ (aligned), otherwise. 
Since human preference is unknown and can only be characterized probabilistically, we multiply this coefficient by the probability $P_o(k)$ from Sec.~\ref{sec:HPM}. $P_o(k)$ is repeatedly evaluated at each time step $k$ using the team's current position as the starting point.
Equation \eqref{eq:phi_evolution} then maps the accumulated discomfort $g(k)$ into human stubbornness \(\phi(k)\) using an exponential function, where \(\phi(0)\in(0,1]\) is the initial stubbornness of the human, both to be determined. Here, the use of the exponential function is due to the fact that humans exhaust their patience more rapidly as their discomfort accumulates, leading them to become stubborn. A similar exponential function has been used to represent changes in human stubbornness in~\cite{heskes2000use, tian2021social}.
\(\eta>0\) is a sensitivity parameter to be identified in our experiments using real human feedback.
{A larger value of $\eta$ yields slower growth in $\phi(k)$ for the same accumulated discomfort, which corresponds to a more tolerant individual.}

\subsubsection{Mode Transition Mechanism} \label{mode_transition_mec}
Depending on human stubbornness, we consider two modes during the co-transportation task. The transition of the modes depends on human stubbornness  \(\phi(k)\):
\begin{itemize}[leftmargin=*]
    \item When $\phi(k) \le 1$, the robot selects a path that actively guides the human.
    \item When $\phi(k) > 1$, the robot follows the human. 
\end{itemize}
At the beginning of the task, since \(\phi(0)\in(0,1]\), the team starts in robot-leading mode. The system transitions into human-leading mode once the stubbornness measure exceeds $1$. In this work, we consider a simple case where the transition occurs only once, i.e., once the human decides to follow their own preference instead of the robot’s guidance, the mode remains unchanged until the task is completed. This assumption aligns with our definition that the $\phi(k)$ in \eqref{eq:phi_evolution} is non-decreasing.

When the system operates in robot-leading mode ($\phi(k) \le 1$), the robot needs to choose a path that accounts for both the objective cost of the robot and the subjective cost of the human. This is formulated as:
\vspace{-0.22cm}
\begin{align}\label{eq_co_path}
    o^*(k) = \arg\min_{o \in \mathcal{O}} \bigl(1-\phi(k)\bigr) \mathcal{J}_{o}(k)
    + \phi(k) \tilde{\mathcal{J}}_{o}(k)
\end{align}

\vspace{-0.4cm}
\noindent
where, 

\vspace{-0.6cm}
\begin{align*}
    \tilde{\mathcal{J}}_{o}(k) = \frac{1}{(\overline{\alpha} - \underline{\alpha})(\overline{\gamma} - \underline{\gamma})} \int_\alpha \int_\gamma \widehat{\mathcal{J}}_{o}(k|\alpha, \gamma) d\alpha \, d\gamma.
\end{align*}

\vspace{-0.15cm}
\noindent
Here, although \eqref{eq_subjc} can evaluate the subject cost of an option $o$ for any given pair of $(\alpha, \gamma)$, the exact human-specific parameters are unknown. Therefore, the human cost for option $o$ must be approximated over all possible parameters, which yields $\tilde{\mathcal{J}}_{o}(k)$. This $\tilde{\mathcal{J}}_{o}(k)$ is also repeatedly evaluated at each time step $k$ using the team's current position as the starting point.
Then in \eqref{eq_co_path}, we determine the path option $o^*(k)$ by weighing the object cost $\mathcal{J}_{o}(k)$ and approximated human subjective cost $\tilde{\mathcal{J}}_{o}(k)$ based on human stubbornness measure \(\phi(k)\). This formulation enables a smooth transition between robot-leading and human-leading modes, because before the transition, \(\phi(k)=1\), meaning that the cost is dominated by the human's subjective cost.
Furthermore, we remark that the path selection problem \eqref{eq_co_path} will not result in locked cycles (i.e., infinite oscillation between options). This is ensured by the monotonic nature of $\phi(k)$, and the fact that once the team selects an option $o^*(k)$, the corresponding costs $\mathcal{J}_{o^*}(k)$ and $\tilde{\mathcal{J}}_{o^*}(k)$ will decrease more than the costs of other options in the next time step.
While we do not provide rigorous proof due to space limitations, this statement is also supported by our experimental observations.

When the system operates in human-leading mode ($\phi(k) > 1$), the robot must anticipate possible options the human may choose and adapt its control accordingly. To achieve this, we incorporate a pose optimization strategy, which is introduced in the following subsection.

\subsection{Pose Optimization in Low-Level Control} \label{PO}

In this subsection, we introduce a pose optimization strategy that improves the low-level control of a mobile manipulator in human-led mode, where the robot must mitigate uncertainty in human choices.
When the human is leading, the robot does not know a priori which path option the human will choose. To mitigate this uncertainty, we exploit the redundancy of the robotic arm and adjust its joint configuration to remain favorable across all potential path choices, weighted by their probabilities of being chosen by the human.

For a given end-effector trajectory, different joint configurations result in varying control costs. To quantify this option-dependent cost, we adopt a receding-horizon Model Predictive Control (MPC) formulation that tracks a nominal trajectory\footnote{The trajectory can be generated by existing algorithms \cite{hart1968formal, warren1993fast}, which plan motion through openings while avoiding obstacles. Details of trajectory generation are omitted as they are not the main contribution of this paper.} associated with each path option $o$. The MPC problem is formulated as:
\vspace{-0.23cm}
\begin{subequations}\label{eq_MPC}
\begin{align}
\min_{u^{(0:H-1)}} \quad 
G_o&=\sum_{t=1}^H \underbrace{\|x(t) - s_{o}(t)\|^2_{Q}}_{\text{tracking cost}}
+ \sum_{t=0}^{H-1} \underbrace{\|u(t)\|^2_{R}}_{\text{control effort}} \\
\text{s.t.} \quad
 &x(t+1) = x(t) + B(q) u(t) \label{eq_dyn}
\end{align}
\end{subequations}
where $t$ is the MPC time step; $H$ is the planning horizon; $s_o(0:H)$ is the nominal trajectory for path option $o$; $Q$ and $R$ are weighting matrices for costs, which are positive semi-definite and positive definite, respectively. 
For a vector $v$, and a positive (semi-) definite matrix $M$, we let $\|v\|_{M}^2 = v^{\top}Mv$.
Equation \eqref{eq_dyn} represents a generic linearized motion dynamics for the mobile manipulator, where $x$ denotes the robot's end-effector state, and $u^{(0:H-1)} = \{ u(0), \cdots, u(H-1) \}$ denotes the control inputs on joints and wheels.  A specific example of this model is used in simulations.
We note that the input matrix $B(q)$ is linearized from the robot's nonlinear dynamics, which depends on the robot's joint configuration, denoted by \(q\). 
Solving \eqref{eq_MPC} yields the optimal cost for option $o$, given the joint configuration $q$, denoted by
$G_o^*(q)\triangleq G_o(u^{*(0:H-1)},q)$.
Note that this is solved conditionally for each option  $o$ to evaluate $G_o^*(q)$.
As the human may choose any options,
control decisions should be informed by the costs across all possible options, weighted by their probability distribution \(P_o\) (cf.\ Section~\ref{sec:HPM}). 
{For that, we select a joint configuration $q$ that minimizes the probability-weighted sum of the option-conditioned optimal MPC costs $\{G_o^*(q)\}_{o\in\mathcal{O}}$ with the probability distribution $P_o$. This yields a scenario-based stochastic formulation where each scenario is a discrete path option. The inner controller remains a standard deterministic MPC for each option's reference trajectory $s_o(t)$. 
Since the main uncertainty we want to model is humans' path choice, not internal or external disturbances, we do not need a stochastic MPC to capture uncertainty in human choices.
For a redundant robotic arm, there are infinitely many feasible poses that track a nominal trajectory. Direct minimization of $G_o(u^{*(0:H-1)},q)$ for optimal $q$ is intractable due to its complex nonlinear and non-analytic dependence on $q$. Instead, we consider a candidate set $\mathcal{Q}$ of feasible joint configurations and solve the following pose optimization problem:

\vspace{-0.45cm}
\begin{equation}
{q}^*
=
\arg\min_{{q}\in\mathcal{Q}}
\Bigl\{
    \sum_{o\in\mathcal{O}}
    P_o G_o(u^{*(0:H-1)},q)
    + \kappa\,\bigl\|q - q_c\bigr\|_2^2
\Bigr\}.
\label{eq:pose_opt}
\end{equation}

\vspace{-0.15cm}
\noindent
{We generate the candidate set $\mathcal{Q}$, as discussed in~\cite{11246189}.}
For each candidate joint configuration $q$, the cost is evaluated across all possible path options $o$, weighted by their probability $P_o$ of human choice. In the second term, we penalize the difference between the current pose $q_c$ and the new pose $q$ for pose adjustment, and $\kappa\in\mathbb{R}_{+}$ is the cost weight.
{This formulation \eqref{eq:pose_opt} differs from standard Jacobian/Inverse-Kinematics singularity-avoidance criteria, which optimize redundancy for a single reference trajectory. Here, the reference depends on the human's uncertain intent.} 
So, by solving for ${q}^*$, the robot proactively prepares for all possible human decisions, 
avoiding poor kinematic configurations, leading to smoother human-robot co-transportation.

\vspace{-0.11cm}
\section{Simulation Results} \label{sec_results}

In this section, we test the three hypotheses stated in Sec.~\ref{sec:problem_interest}. We use two human-feedback studies to (i) fit and validate the 
human preference uncertainty model
and (ii) estimate the time-varying stubbornness sensitivity parameter. 
We then use simulation to evaluate performance under human intent uncertainty with the pose optimization module.

\vspace{-0.20cm}
\subsection{\textbf{Hypothesis tested: H1.} \textit{If human preference parameters are uncertain rather than fixed, then a probabilistic model over options can better represent how humans distribute their choices across multiple paths.}}

\begin{figure} [t]
    \centering
    \includegraphics[width=0.38\textwidth]{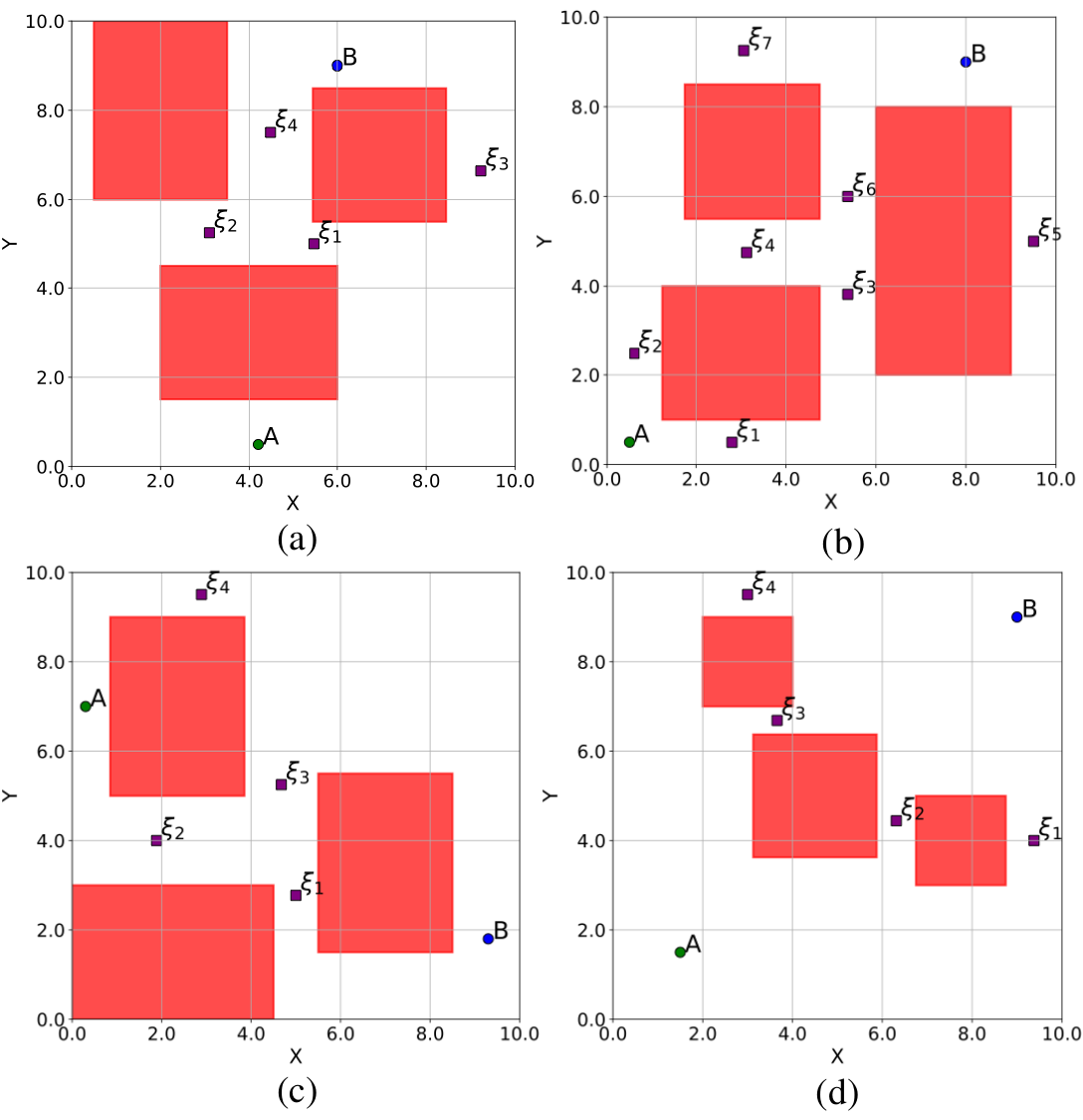}
    \caption{\small  Start position $A$, target position $B$. 
    (a) \textbf{Environment 1:} Options: \{$(A, \xi_2, \xi_1, \xi_3, B)$, $(A, \xi_2, \xi_4, B)$, $(A, \xi_1, \xi_4, B)$, $(A, \xi_3, B)$\}. 
    (b) \textbf{Environment 2:} Options: \{$(A, \xi_2, \xi_4, \xi_6, B)$, $(A, \xi_1, \xi_3, \xi_6, B)$, $(A, \xi_2, \xi_7, B)$,  $(A, \xi_1, \xi_5, B)$.\} 
    (c) \textbf{Environment 3:} Options: \{$(A, \xi_4, B)$, $(A, \xi_2, \xi_1, B)$, $(A, \xi_2, \xi_3, B)$,  $(A, \xi_4, \xi_3, \xi_1, B)$.\} 
    (d) \textbf{Environment 4:} Options: \{$(A, \xi_4, B)$, $(A, \xi_3, B)$,  $(A, \xi_2, B)$, $(A, \xi_1, B)$\}.
    We use this figure to collect human responses, where the trajectories through openings are not visualized so that participants do not get any extra information.
    \vspace{-2.3em}
    }
    \label{exp_environemtns}
\end{figure}


For the experiment, following our model description, we set $\alpha\sim\mathrm{U}(0.55, 0.95)$ and $\gamma\sim\mathrm{U}(0.5, 1.1)$, which captures the range of results from existing literature~\cite{tversky1992advances,prelec1998probability, abdellaoui2007loss, thomas2023psychophysics} through extensive human experiments. To implement human probability model~\eqref{eq_Po}, we first need to determine $c_{\text{r}}$ for human subjective cost~\eqref{eq_subjc}. 

\noindent\textbf{Experiment design:} We create four different environments with different combinations of obstacles and start-target positions, as shown in Fig.~\ref{exp_environemtns}. The robot's diameter is set as $d_{\text{r}} = 0.5\,\text{m}$.  In each environment, 
we consider four reasonable path options to reach the target through different openings, each with varying risk levels and distances. 
We show these environments to twenty human participants and obtain their feedback on how likely they would choose each option in percentages. Participants use visual observation and do not know the objective evaluation of risks and distances. 
We instructed the participants to allocate probabilities that sum to $100\%$ across the four options.
To determine the $c_{\text{r}}$ values, we fit the model using the average percentage of the feedback from environments 1-3. The resulting value is $c_{\text{r}} = 10$. The model-fitting column of Table~\ref{Tab_probability_distribuion} presents the average percentages provided by humans alongside the corresponding model output (in parentheses) for each option in these three environments.

\noindent\textbf{Model validation:} We then validate this value by comparing the model's output with the human responses in the fourth environment, where we observe a close alignment in each option. 
Across all environments, we observe that our model can capture the ranking of options in terms of how frequently that option may be selected.
Additionally, for low-probability choices, where humans assigned a small probability to less favorable paths, the model often rounded these probabilities down to zero. This suggests that while the model effectively identifies the most favorable options, it may overlook certain outlier choices. Overall, we found that humans tend to assign higher percentages to relatively safer options, even when this results in longer travel distances. 
These results support \textbf{H1} as the predicted choice distribution by the proposed model matches the human ranking of options across environments.

\begin{table}[t]
\centering
\caption{\small Probability Distribution of Human Choices (\%). Proposed model outputs are in parentheses.
}
\resizebox{0.39\textwidth}{!}{
\begin{tabular}
{|c|c|c|c|c|}
\hline
\multirow{2}{*}{} 
  & \multicolumn{3}{c|}{\textbf{Model-Fitting}}
  & \multicolumn{1}{c|}{\textbf{Validation}}
  \\ \cline{2-5} 
 & \textbf{Env 1} & \textbf{Env 2} & \textbf{Env 3} & \textbf{Env 4} \\ \hline
\textbf{Option 1} & 5 (0) & 45 (51.6) & 12.5 (5.1) & 8 (3.6) \\ \hline
\textbf{Option 2} & 45 (39.5) & 13.5 (8.7) & 30 (35.5) & 15 (18.3) \\ \hline
\textbf{Option 3} & 40 (36.1) & 6.5 (0) & 47.5 (59.4) & 41 (46.3) \\ \hline
\textbf{Option 4} & 10 (24.4) & 35 (39.7) & 5 (0) & 36 (31.8) \\ \hline
\end{tabular}
}
\label{Tab_probability_distribuion}
\vspace{-2.0em}
\end{table}


\vspace{-0.15cm}
\subsection{\textbf{Hypothesis tested: H2.} \textit{If the robot uses a time-varying stubbornness measure to estimate when the human would take the lead, then the team can achieve a lower overall objective cost than a fixed-mode baseline.}}
\vspace{-0.15cm}
Here, we evaluate the mode-switching mechanism described in Sec.~\ref{HBM}. We start by identifying the value $\eta$ for different participants to implement the stubbornness measure~\eqref{eq:phi_evolution}, then incorporate it into the mode transition.

\noindent\textbf{Experiment design:} We use the environments shown in Fig.~\ref{exp_environemtns}. For each environment, we create videos of progressively moving trajectories for each path option, representing a robot leading the movement.
We show these videos to participants and, during the process, ask them if they want to stop following the robot and switch to another option. If they choose to do so, we record
that feedback time.
This time corresponds to the mode transition, i.e., $\phi(k)$ exceeds 1. We use this time and the trajectory to calculate the individual $\eta$ for twenty participants.
Among the twenty $\eta$, we select three representative ones, corresponding to sensitive ($\eta=5$), medium ($\eta=10.5$), and less sensitive ($\eta=15$) individuals to present the following results.

\noindent\textbf{Model validation:} 
To quantitatively evaluate our model, we compare the robot's objective costs \eqref{eq:robot_utility} with and without the mode transition mechanism across all four environments. For each environment, we randomly sample twenty different ($\alpha$, $\gamma$) values from their respective ranges and compute the average accumulated objective cost with standard deviation for the resulting 
human-robot 
trajectories, 
presented in Table~\ref{Tab_average_objective_costs}.
This table shows that across all environments, using the mode transition model with different $\eta$ values consistently yields lower objective costs for the team than fixed human-leading mode, denoted as ``Without Model''. In this baseline, the robot follows the human preferred option obtained by~\eqref{eq_H_decision} from start to end, without depending on $\eta$. 
When $\eta=5$, the mode transition happens early, resulting in slightly lower objective costs than without
the model. With $\eta = 10.5$, the model yields similar costs in environments 1 and 2 but achieves greater cost reduction in environments 3 and 4. This reflects that by the time the human stubbornness parameter exceeded 1, the human started to prefer the option initially favored by the robot for the team. Furthermore, when $\eta=15$, 
the robot successfully guides the human to adopt its initially selected path
in all environments. 
These results statistically and quantitatively support \textbf{H2} and verify the effectiveness of our model in guiding the team toward safer and more efficient paths
with
defined objective cost (~\ref{eq:robot_utility}).
It is also worth noting that in all experiments, we do not observe locked cycles when the robot is leading, as discussed in Sec.~\ref{HBM}.

\begin{table}[t]
\centering
\caption{{\small Average objective cost under different human stubbornness sensitivity (mean$\pm$std).}}
\label{Tab_average_objective_costs}
\scriptsize
\setlength{\tabcolsep}{3pt}
\renewcommand{\arraystretch}{1.10}
\resizebox{0.41\textwidth}{!}{
\begin{tabular}{|
>{\arraybackslash}m{1.65cm}|
>{\centering\arraybackslash}m{1.15cm}|
>{\centering\arraybackslash}m{1.15cm}|
>{\centering\arraybackslash}m{1.15cm}|
>{\centering\arraybackslash}m{1.15cm}|
}
\hline
\textbf{Method} & \textbf{Env 1} & \textbf{Env 2} & \textbf{Env 3} & \textbf{Env 4} \\
\hline
\textbf{Without Model} 
& $1939.67\!\pm\!38.6$ & $1962.98\!\pm\!44.3$ & $1946.48\!\pm\!41.2$ & $2279.98\!\pm\!52.3$ \\
\hline
\textbf{With $\eta=5$}   
& $1925.38\!\pm\!32.1$ & $1895.16\!\pm\!28.7$ & $1927.35\!\pm\!30.5$ & $2213.82\!\pm\!41.3$ \\
\hline
\textbf{With $\eta=10.5$}   
& $1916.11\!\pm\!30.6$ & $1887.75\!\pm\!27.9$ & $1853.84\!\pm\!26.4$ & $1767.07\!\pm\!24.8$ \\
\hline
\textbf{With $\eta=15$}   
& $1755.42\!\pm\!23.5$ & $1586.56\!\pm\!20.4$ & $1826.19\!\pm\!25.9$ & $1767.07\!\pm\!24.8$ \\
\hline
\end{tabular}
}
\vspace{-2.95em}
\end{table}

 \begin{figure} [b]
    \vspace{-1.9em}
    \centering
    \includegraphics[width=0.48\textwidth]{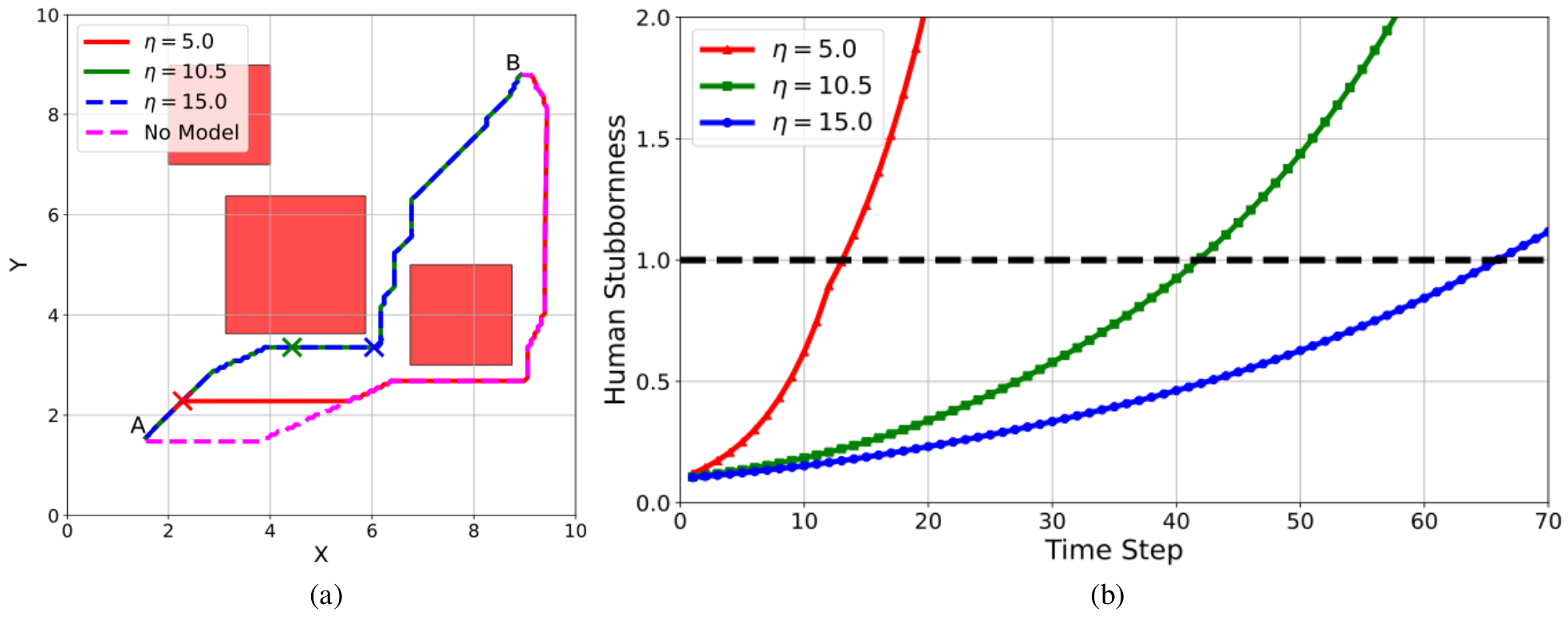}
    \caption{\small (a) Trajectories the human-robot team followed during the co-transportation task with $\eta=5, 10.5, 15$, and the trajectory with no mode transition mechanism. The cross ($\times$) marks represent the moment in which the stubbornness parameter exceeds 1, (b) Change in stubbornness over time with different $\eta$ values.}
    \label{fig_human_stubbornness}
\end{figure}

In addition to cost comparison, we provide a specific example from Environment 4 to illustrate how mutual adaptation enables the robot to influence human decisions and improve the team's behavior. Specifically, in this scenario, the robot prefers path option 3 $(A, \xi_2, B)$ based on objective evaluation, whereas the human subjectively prefers option 4 $(A, \xi_1, B)$ (with the highest probability).
We showcase the resulting team trajectory under four cases, assuming the human has three different stubbornness sensitivity $\eta=5, 10.5, 15$, respectively, or when the mode transition mechanism is not applied (i.e., the fixed human-leading mode).
The corresponding trajectories are shown in Fig.~\ref{fig_human_stubbornness}-a; the evolution of human stubbornness over time is shown in Fig.~\ref{fig_human_stubbornness}-b. 
For $\eta=5$, the human becomes stubborn quickly, which limits the robot's influence on the team. Consequently, their trajectory almost follows the case without the model transition mechanism, i.e., they follow the human preferred path option 4. 
On the other hand, if $\eta=10.5$ or $15$, we observed that at a later time (time steps 42, 66, respectively), $\phi(k)$ exceeds 1, and the robot transitioned to adapting to the human.
However, at this moment, the team’s current position leads the human to change their preference to option 3, as shown in Fig.~\ref{fig_human_stubbornness}-a. Thus, the team completes the task using this path. 
This demonstrates that the robot successfully influences human behaviors before they become fully stubborn, ultimately reducing the team's cost. These results verify the effectiveness of our model. With a gradually increasing human stubbornness parameter and the robot’s mode transition mechanism, the team completes the task using a more efficient path.
Thus, this example qualitatively supports \textbf{H2}.

\vspace{-0.20cm}
\subsection{\textbf{Hypothesis tested: H3.} \textit{If the robot performs pose optimization in human-leading mode to prepare for multiple possible human choices, then the robot can reduce low-level control cost compared to not using pose optimization.}}

We evaluate the effectiveness of our pose optimization (PO) model to better compensate for varying human uncertainties described in Sec.~\ref{PO} through simulation.

\noindent\textbf{Experiment Design:} 
To obtain the optimized joint configuration $q^*$, we generate a set of candidate joint configurations $\mathcal{Q}$ with $|\mathcal{Q}|=15$. 
For MPC, we set planning horizon $H=8$, weight matrices $Q = 1000\, I_6$ and $R = I_9$, and choose $\kappa = 1$ for pose optimization.
To simulate uncertainties, we randomly create four ranges of human preference parameters ($\alpha\pm \omega_\alpha$, $\gamma\pm \omega_\gamma$) defined as sub-ranges of $\alpha\sim\mathrm{U}(0.55, 0.95)$ and $\gamma\sim\mathrm{U}(0.5, 1.1)$. These parameters may change once or twice during the task, representing shifts in human preferences. The robot utilizes pose optimization to adapt to this.

\noindent\textbf{Model validation:} 
To validate our model, we define the system's overall true control cost as:

\vspace{-0.45cm}
\begin{align*}
    \mathcal{C} = \sum_{t=1}^T \|x(t) - s(t)\|_Q^2 + \|u(t)\|_R^2 + \kappa \|q-q_c\|^2_2
\end{align*}

\vspace{-0.15cm}
\noindent
This equation accumulates true cost over time by considering tracking error, control input, and pose optimization, where $s$ represents the true human state in simulation. 
The robot dynamics follow the form of \eqref{eq_dyn}. 
The result is then compared with a baseline where we do not consider pose optimization. For each human in each environment, we conduct 100 random experiments, with the average costs and standard deviation presented in Table~\ref{Tab_average_pose_optimization}. 
As the table indicates, incorporating pose optimization in human-led mode reduces the system's true cost in each environment compared to when pose optimization is not used. Notably, although \eqref{eq:pose_opt} is formulated based on the probability distribution of human choices (i.e., using the overall range of $\alpha$ and $\gamma$, rather than the sub-ranges, which are assumed to be unknown to robots), the resulting optimized joint configuration proves beneficial when collaborating humans with varying preference parameters.
These results support \textbf{H3} by showing lower control cost under human intent uncertainty when PO is enabled.

\begin{table}[t]
\centering
\caption{\small Total robot control cost with or without pose optimization (PO) (mean$\pm$std).
Settings: S1$(\alpha{=}0.65\pm0.1,\gamma{=}0.6\pm0.1)$,
S2$(0.75\pm0.2,0.7\pm0.2)$,
S3$(0.8\pm0.15,0.85\pm0.15)$,
S4$(0.85\pm0.1,0.95\pm0.15)$. For $\alpha$ and $\gamma$, $\pm$ represents parameter range.}
\label{Tab_average_pose_optimization}

\scriptsize
\setlength{\tabcolsep}{3pt}
\renewcommand{\arraystretch}{1.10}
\resizebox{0.48\textwidth}{!}{

\begin{tabular}{|
>{\centering\arraybackslash}m{0.8cm}|
>{\centering\arraybackslash}m{0.65cm}|
>{\centering\arraybackslash}m{1.55cm}|
>{\centering\arraybackslash}m{1.55cm}|
>{\centering\arraybackslash}m{1.55cm}|
>{\centering\arraybackslash}m{1.55cm}|
}
\hline
\textbf{} & \textbf{Env} & \textbf{S1} & \textbf{S2} & \textbf{S3} & \textbf{S4} \\
\hline

\multirow{4}{*}{\makecell[c]{\textbf{With}\\\textbf{PO}}}
& \textbf{Env 1} & $843.9\pm18.2$  & $781.5\pm16.1$  & $1546.5\pm31.8$ & $805.3\pm17.1$ \\
\cline{2-6}
& \textbf{Env 2} & $1267.7\pm26.4$ & $1094.4\pm23.1$ & $1130.1\pm24.6$ & $1162.2\pm25.2$ \\
\cline{2-6}
& \textbf{Env 3} & $705.2\pm14.9$  & $639.3\pm13.8$  & $1003.8\pm21.4$ & $796.2\pm16.5$ \\
\cline{2-6}
& \textbf{Env 4} & $760.2\pm15.7$  & $1424.4\pm29.8$ & $822.1\pm17.1$  & $972.9\pm20.1$ \\
\hline

\multirow{4}{*}{\makecell[c]{\textbf{Without}\\\textbf{PO}}}
& \textbf{Env 1} & $1078.3\pm33.5$ & $992.7\pm31.2$  & $1759.1\pm52.1$ & $1315.1\pm41.8$ \\
\cline{2-6}
& \textbf{Env 2} & $2139.7\pm67.8$ & $1688.1\pm55.4$ & $1180.1\pm36.9$ & $1281.6\pm39.7$ \\
\cline{2-6}
& \textbf{Env 3} & $732.8\pm22.6$  & $647.1\pm20.3$  & $1200.4\pm37.8$ & $1034.3\pm33.1$ \\
\cline{2-6}
& \textbf{Env 4} & $890.9\pm28.1$  & $2332.5\pm78.4$ & $1136.6\pm35.7$ & $981.2\pm30.8$ \\
\hline
\end{tabular}
}
\vspace{-2.5em}
\end{table}

\section{Conclusion}


In this paper, we introduced a hypothesis-driven framework to address two main challenges in human-robot co-transportation without direct communication:  the uncertainty of human preference parameters and the need to balance adaptation strategies that benefit both humans and
robots. We derived a model that captures the probability distribution of human choices rather than relying on fixed human preference parameters. We then integrated a time-varying stubbornness measure to analyze the mode transition from robot-leading to human-leading mode. Following these, we proposed a pose optimization strategy that selects redundant joint configurations to reduce low-level control costs under human intent uncertainty when the human is leading. We validated our framework through studies with human participants.
We used human feedback to fit and validate the probability distribution model 
and to estimate the time-varying stubbornness sensitivity parameter used in the mode transition. Then, we used simulations to demonstrate reduced objective cost with the mode transition mechanism and reduced control cost with pose optimization under human intent uncertainty. These results support our hypotheses \textbf{H1}, \textbf{H2}, and \textbf{H3}.
A limitation of our approach is that we use static images and videos to collect human feedback. This does not capture the full physical interaction of co-transportation.
Future work will include physical human-robot co-transportation studies, additional human factors (e.g., fatigue, stress), and extend to multi-human multi-robot co-transportation scenarios.


\bibliography{bibliography}
\bibliographystyle{ieeetr}

\end{document}